\documentclass[final]{cvpr}

\usepackage{times}
\usepackage{epsfig}
\usepackage{graphicx}
\usepackage{amsmath}
\usepackage{amssymb}
\usepackage{booktabs}
\usepackage{caption}
\usepackage{unitsdef}
\usepackage{tabu}
\usepackage{multirow}
\usepackage[dvipsnames]{xcolor}
\usepackage{subcaption}

\usepackage[pagebackref=true,breaklinks=true,colorlinks,bookmarks=false]{hyperref}
\usepackage[capitalize]{cleveref}
\crefname{section}{Sec.}{Secs.}
\Crefname{section}{Section}{Sections}
\Crefname{table}{Table}{Tables}
\crefname{table}{Tab.}{Tabs.}

\makeatletter
\newcommand*\bigcdot{\mathpalette\bigcdot@{.5}}
\newcommand*\bigcdot@[2]{\mathbin{\vcenter{\hbox{\scalebox{#2}{$\m@th#1\bullet$}}}}}
\makeatother

\def\cvprPaperID{5409} % *** Enter the CVPR Paper ID here
\def\confYear{CVPR 2022}

\begin{document}

\title{Depth-Guided Sparse Structure-from-Motion for Movies and TV Shows}
%\title{Structure-from-Motion for Movies and TV series Guided by Depth-Priors}

%\teaser{
%	\vspace{0.1cm}
%	\includegraphics[width=1.0\textwidth]{figures/teaser.jpg}
%	\vspace{-0.6cm}\captionof{figure}{\small {\textbf{Approach Overview -- } An example shot from a TV episode with small motion-parallax is illustrated. Standard geometry-based SfM approaches (\textit{e.g.}, COLMAP~\cite{colmap2016}) do not work for such small motion-parallax videos, as illustrated in column-$2$ by the wrongly inferred scene-structure (mostly uniform depth) and the incorrectly estimated camera-path (vertical). To address this challenge, we propose a simple yet effective approach that leverages single-frame based depth-prior obtained from a pre-trained network (DPT-large~\cite{dptlarge2021}) to significantly improve the prevalent SfM pipeline of COLMAP~\cite{colmap2016} applied to our small motion-parallax setting of movies and TV shows.
%	}}\vspace{-0.25cm}
%\label{fig:teaser_fig}
%}
\author{Sheng Liu\thanks{This work was done when the author was an intern at Amazon.}\\
University at Buffalo\\
{\tt\small sliu66@buffalo.edu }
\and
Xiaohan Nie, {} Raffay Hamid\\
Amazon Prime Video\\
{\tt\small \{nxiaohan, raffay\}@amazon.com}
}

%To address this challenge, we leverage single-image depth-prior obtained from a pre-trained network to significantly improve the prevalent incremental SfM pipeline of COLMAP~\cite{colmap2016} applied small-parallax setting of movies and TV shows.

\maketitle

\begin{abstract}

\noindent Existing approaches for Structure from Motion (SfM) produce impressive $3$-D reconstruction results especially when using imagery captured with large parallax. However, to create engaging video-content in movies and TV shows, the amount by which a camera can be moved while filming a particular shot is often limited. The resulting small-motion parallax between video frames makes standard geometry-based SfM approaches not as effective for movies and TV shows. To address this challenge, we propose a simple yet effective approach that uses single-frame depth-prior obtained from a pretrained network to significantly improve geometry-based SfM for our small-parallax setting. To this end, we first use the depth-estimates of the detected keypoints to reconstruct the point cloud and camera-pose for initial two-view reconstruction. We then perform depth-regularized optimization to register new images and triangulate the new points during incremental reconstruction. To comprehensively evaluate our approach, we introduce a new dataset (\textbf{StudioSfM}) consisting of $130$ shots with $21\textrm{K}$ frames from $15$ studio-produced videos that are manually annotated by a professional CG studio. We demonstrate that our approach: (a) significantly improves the quality of $3$-D reconstruction for our small-parallax setting, (b) does not cause any degradation for data with large-parallax, and (c) maintains the generalizability and scalability of geometry-based sparse SfM. Our dataset can be obtained at \href{https://github.com/amazon-research/small-baseline-camera-tracking}{https://github.com/amazon-research/small-baseline-camera-tracking}.

\end{abstract}\vspace{-0.5cm}
\section{Introduction}
\label{sec:intro}

\begin{figure}[!t]
	\centering
	\includegraphics[width=0.30\textwidth]{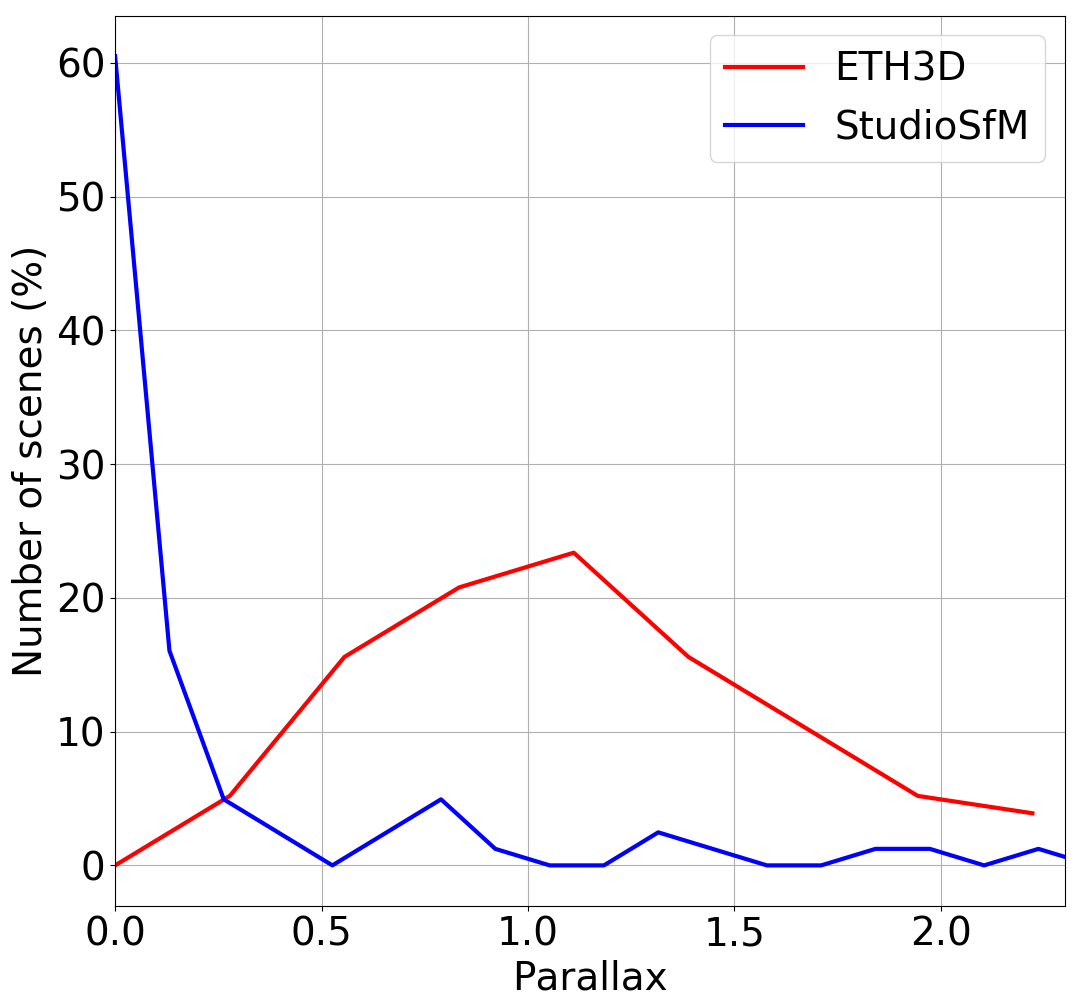}
\vspace{-0.2cm} \caption{\small {\textbf{Comparison of Parallax-Distribution:} Parallax-distribution of ETH3D~\cite{eth3d2017} is plotted with StudioSfM -- a new dataset with $21\textrm{K}$ frames in $130$ manually annotated shots from $15$ TV episodes (see $\S$~\ref{ss:datasets} for details of data and computation of parallax). The long-tail distribution of StudioSfM shows that small-motion parallax is significantly more common in studio-produced content than in standard SfM datasets.}}
	\label{fig:parallax}
	\vspace{-0.3cm} 
\end{figure}

\noindent Estimating camera motion and $3$-D scene geometry in movies and TV shows is a standard task in video production. Existing Structure from Motion (SfM) approaches for $3$-D scene reconstruction produce high-quality results especially for images with large parallax~\cite{wide2020, dai2017scannet, nyu2012, eth3d2017}. However, creating engaging viewing experience in movies and TV shows often constrains the amount of camera movement while filming a shot. This often leads to insufficient parallax compared to standard SfM datasets captured specifically for $3$-D reconstruction (see Figure~\ref{fig:parallax} for more details).

%This often leads to insufficient parallax (see Figure~\ref{fig:parallax} for more details), which limits the effectiveness of using geometry-based SfM approaches~\cite{review2015} to extract camera pose and geometry in movies and TV shows.

\begin{figure*}
  \centering
  \includegraphics[trim={0, 0pt, 0pt, 0pt}, clip, width=0.85\linewidth]{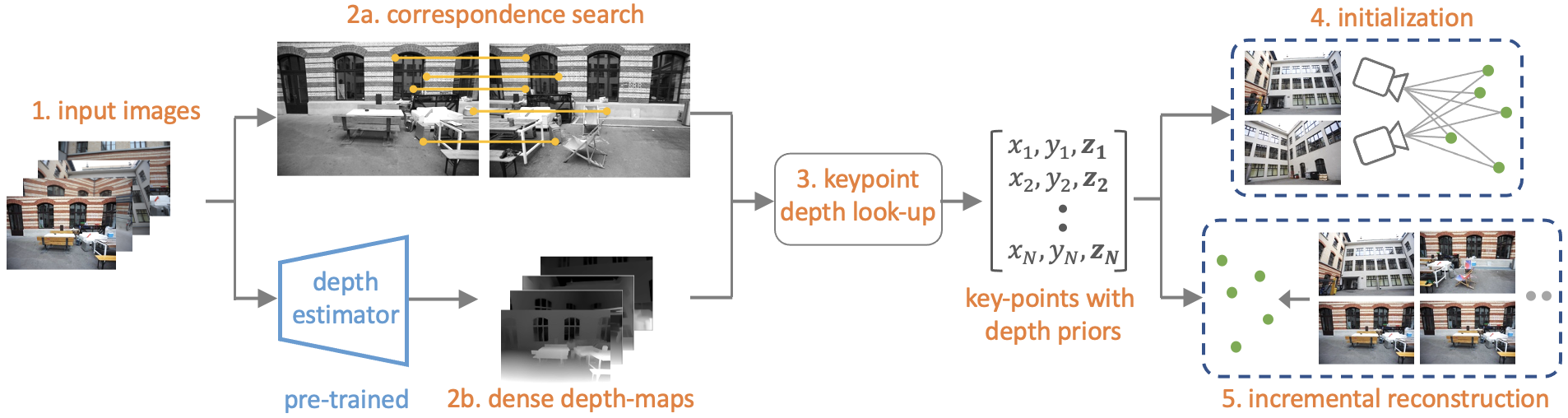}
  \caption{\small{\textbf{Proposed Pipeline -- } Given a set of input images (Step $1$), we detect $2$-D keypoints and match them across frames, \textit{i.e.}, correspondence search (Step $2$-a), as well as use a pretrained network to obtain their single-frame based dense depth-maps (Step $2$-b). We apply bi-linear interpolation to look-up the depths of the detected key-points from these dense depth-maps (Step $3$). We use the detected 2-D keypoints along with their depth-priors to improve the initialization (Step $4$) and incremental reconstruction steps (Step $5$).}}
  \vspace{-0.3cm}
  \label{fig:pipeline}
\end{figure*}

%Specifically, this challenge arrises as geometry-based SfM approaches~\cite{pmvs2010, rome2010, colmap2016, opensfm2019, globalsfm2014, openmvg2016} first detect $2$-D image keypoints~\cite{lowe_2004, harris_1988, orb_2011, super2018} followed by matching them across multiple frames and recovering the $3$-D geometry and camera motion based on motion-parallax. 
This insufficient parallax is one of the key challenges~\cite{review2015} that limits the effectiveness of well-developed geometry-based SfM approaches~\cite{pmvs2010, rome2010, colmap2016, opensfm2019, globalsfm2014, openmvg2016} that recover camera motion and geometry based on the principle of motion-parallax. Shots with small motion-parallax are ill-conditioned for $3$-D reconstruction as algebraic methods for two-view reconstruction are numerically unstable in such situations~\cite{invitation2005}. Conventional SfM pipelines (\textit{e.g.}, COLMAP~\cite{colmap2016}) use various heuristics to handle small-parallax data, \textit{e.g.}, by using inlier ratio to decide the two-view motion type to prevent two-view reconstruction from using panoramic image pairs, and filtering out points with small triangulation angles. These heuristics however require careful tuning and can fail completely when using data which has no image pairs with sufficient parallax.

In contrast, learning-based approaches~\cite{posenet2015, egodepth2017, demon2017, deepsfm2020} are able to handle data with small parallax more effectively as they can learn to predict depth and pose from large-scale labelled datasets. However, as these methods do not incorporate geometric-consistency constraints between images, their pose and depth estimates are not as accurate~\cite{rcvd2021}. Furthermore, the generalizability of these approaches heavily depends on the scale of labeled data used for their training, which can be laborious and expensive to collect.

Recently proposed hybrid approaches~\cite{banet2019, deepv2d2020, rcvd2021} have achieved more accurate results than learning-based approaches by employing learned depth priors as implicit constraints for geometric consistency. However, these approaches do not use robust estimators thus making them heavily dependent on the quality of used optical flow which can adversely affect their robustness. Moreover, these approaches require heavy compute and memory resources. This prevents them from scaling to larger problems.

\vspace{0.05cm} \noindent {\textbf{Key Contributions:}} To address these challenges, we propose a novel hybrid approach that combines the strengths of: (a) geometry-based SfM to achieve high-accuracy without requiring additional labelled data, and (b) learning-based SfM to effectively handle data with insufficient parallax. As illustrated in Figure~\ref{fig:pipeline}, our approach builds on the standard geometry-based SfM pipeline and particularly improves its initialization and incremental reconstruction steps by leveraging single-frame depth-priors obtained from a pretrained deep network. Specifically:

%previously learned monocular depth-priors. Specifically:

% while keeping the accuracy and robustness of its carefully designed and well-engineered modules. Specifically:

\vspace{0.05cm} \noindent $\bullet$ Instead of using epipolar geometry for initial two-view reconstruction, we directly utilize monocular depth obtained from a pretrained model to accurately recover the initial camera pose and point cloud.

\vspace{0.05cm} \noindent $\bullet$ During the incremental reconstruction step, we propose a depth-prior regularized objective function to be able to accurately register and triangulate new images and points.

\vspace{0.1cm} \noindent We demonstrate that our approach is robust to a variety of pretrained networks used to obtain the depth-prior, and maintains the generalizability and scalability of geometry-based SfM pipeline by maximally relying on its well-engineered implementations (\textit{e.g.} COLMAP~\cite{colmap2016}).

To comprehensively evaluate our approach, we collect a new dataset (\textbf{StudioSfM}) containing $130$ shots with $21\textrm{K}$ frames from $15$ TV-episodes. The ground truth camera pose and point clouds were created manually by professionals using commercial CG software (see $\S$~\ref{ss:datasets} for details). We use StudioSfM to demonstrate that our approach offers significantly more accurate camera poses and scene geometry over existing state-of-the-art approaches under small-parallax setting in studio-produced content, while does not cause any degradation on standard SfM datasets~\cite{eth3d2017} with large parallax, and maintains the generalizability and scalability of standard SfM pipelines.
%(\textit{e.g.}, COLMAP~\cite{colmap2016})

%In order to have a comprehensive evaluation on our approach, we collect a new dataset \textbf{StudioSFM} which consists of 133 video clips from 15 studio-level movies and TV episodes. The ground truth camera pose and point cloud are created manually by experts using commercial visual effects software. StudioSfM has much more diverse parallax than existing standard SfM datasets which only contain data with large parallax,  \iffalse The experiments on both StudioSFM and ETH3D demonstrate that our approach drastically improve the accuracy of camera pose and scene geometry using various off-the-shelf pretrained depth estimators for the small parallax data over other approaches, and maintain the same high accuracy and computation efficiency for large parallax data as COLMAP. \fi

%The contributions of our work are summarized below:
%\begin{itemize}
%\item we introduce a novel approach to utilize the single-image depth to improve the initialization and incremental reconstruction for incremental SfM.
%\item our approach does not need ground truth data for training and shows strong performance with a diverse set of off-the-shelf pretrained depth estimators.
%\item we collect a new dataset from movies and TV series and the experiments demonstrate that our approach significantly improves the 3D reconstruction quality for small parallax data over previous literature and does not regress for large parallax data.
%\end{itemize}

\section{Related Work}
\label{sec:related_work}
\noindent \textbf{a. Geometry-Based SfM:} Geometry-based SfM~\cite{rome2010, colmap2016, opensfm2019, globalsfm2014, openmvg2016} approaches have undergone tremendous improvements over the past few decades in terms of their robustness, accuracy, completeness and scalability. Most of these approaches first detect and match local image features~\cite{lowe_2004, harris_1988, orb_2011, super2018}, followed by estimating the two-view motion using epipolar geometry~\cite{dlt_2004} and then reconstructing the $3$-D scene either globally or incrementally using bundle adjustment~\cite{ba2000}. One of the most widely used open-source geometry-based SfM pipeline is COLMAP~\cite{colmap2016} which is often used as a preliminary step for state-of-the-art dense reconstruction approaches~\cite{cvd2020, megadepth2018, nerf2020}. Like most geometry-based approaches however, it requires images with sufficiently large baselines. Our approach improves COLMAP~\cite{colmap2016} to make it work robustly for small parallax setting often found in movies and TV shows.

Previous geometry-based SfM approaches geared for videos with small motion~\cite{accidental2014} simplify the rotation matrix and parameterize bundle adjustment using inverse depth of reference image. Work in~\cite{small2016} makes the same simplification and parameterization as~\cite{accidental2014} but assumes that camera intrinsics are unknown and optimizes them in bundle adjustment. These works show improved results only for videos with very small accidental motion and do not generalize to data with relatively larger motion as is the case in movies and TV shows. Unlike~\cite{accidental2014, small2016} that use priors for camera motion, our approach uses priors for scene geometry which is robust to both narrow as well as wide baseline data.

%Work in \cite{accidental2014} targets videos with accidental small motion by simplifying the rotation matrix and parameterizing the bundle adjustment using inverse depth of reference image. Similarly, work in~\cite{small2016} makes the same simplification and parameterization as~\cite{accidental2014} but assumes that camera intrinsics are unknown and optimizes it in bundle adjustment. These works show improved results only for videos with accidental motion and do not generalize to data with relatively larger motion. Unlike~\cite{accidental2014, small2016} that use priors for camera motion, our approach uses priors for scene geometry which is robust to both narrow as well as wide baseline data.

\vspace{0.1cm} \noindent \textbf{b. Learning-Based SfM:} To jointly estimate motion and depth in an end-to-end fashion, work in~\cite{demon2017} stack multiple encoder-decoder networks for their iterative estimation. To improve the robustness of pose estimation, work in~\cite{deepsfm2020} construct pose-cost volume similar to depth-cost volume~\cite{costvolume2019} used in stereo matching to predict camera pose iteratively. Unlike~\cite{demon2017, deepsfm2020} which rely on ground truth labels for training, our approach utilizes off-the-shelf pre-trained depth estimators without the need of labels from target data.

\vspace{0.1cm} \noindent \textbf{c. Hybrid SfM:} Hybrid approaches attempt to optimize camera pose and depth by using geometry-consistency constraints. Work in~\cite{banet2019} represents depth as a linear combination of depth basis maps, and computes the camera motion and depth by aligning deep features using differentiable gradient descent. Work in~\cite{deepv2d2020} uses dense optical flow to build dense correspondences, and iterates between learning based depth estimation and optimization based motion estimation. Work in~\cite{rcvd2021} optimizes the re-projection loss by allowing depth to deform as splines for low-frequency alignment. Depth filters are used for high-frequency alignment to recover the details. In our approach, we do not rely on optical flow which enables our approach to work on both videos as well as un-ordered image-sets.

\vspace{0.1cm} \noindent \textbf{d. Monocular Depth Estimation:} Recent improvements in deep networks and the availability of large-scale depth-data have contributed to the remarkable progress in monocular depth-estimation~\cite{khan2020deep}. Work in~\cite{monodepth2019} learns a depth estimation network in a self-supervised manner using monocular videos. Work in~\cite{midas2020} focuses on mixing multiple datasets for training using multiple objectives which are invariant to depth scale and range. Work in~\cite{adabin2021} divides depth into bins whose centers are estimated adaptively per-image, and are linearly combined to predict the final depth value. We use off-the-shelf pre-trained monocular depth estimators to generate depth-priors for sparse keypoints. Although monocular depth estimates are inconsistent across frames, we show that using them as priors in SfM pipeline helps the reconstruction process to converge to a better solution.

\section{Approach}
\label{sec:approach}

\subsection{Review of Incremental SfM}
\label{ss:inc_sfm}
\noindent As our approach builds on incremental SfM, for completeness we first review the standard incremental SfM pipeline~\cite{colmap2016} which can be roughly divided into three key components: (i) correspondence search, (ii) initialization, and (iii) incremental reconstruction. We provide details of these component in the following.

\vspace{0.1cm} \noindent \textbf{a. Correspondence Search:} For each image $\textbf{I}$ in a given set of $\textrm{N}$ images $\mathcal{I}$, their $2$-D keypoints $\mathbf{p} \in \mathbb{R}^2$ and respective appearance-based descriptors are extracted and used to match all image-pairs $(\textrm{I}_a, \textrm{I}_b)$ $\in$ $\mathcal{I}$ using a similarity-metric based on their keypoint-descriptors. A robust estimator such as RANSAC~\cite{ransac1981} is used to perform robust geometric verification of the matched image-pairs in order to estimate the geometric transformation between them.

\noindent \textbf{b. Initialization:} Based on epipolar geometry of the corresponding $2$-D keypoints in a matched image-pair ($\mathbf{I}_a, \mathbf{I}_b$), two-view reconstruction is performed to estimate the initial camera pose $(\mathbf{R}_{\textrm{init}}, \mathbf{t}_{\textrm{init}}) \in \mathrm{SE}(3)$ and $3$-D point cloud $\mathbf{P} \in \mathbb{R}^{3}$. Recall that good initialization is critical in incremental SfM pipelines as later steps may not be able to recover from a poor initialization. To this end, heuristics such as number of keypoint matches, triangulation angles and geometric-transformation types are used to select a good image-pair likely to result in high-quality initialization~\cite{colmap2016}.

\noindent \textbf{c. Incremental Reconstruction:} New images from the remaining image-set are incrementally incorporated into the reconstruction process by iterating between the following three steps. $\mathsf{i-Image \ Registration:}$ this step registers a new image to the current $3$-D scene by first solving the Perspective-n-Point (PnP) problem~\cite{andrew2001multiple} using RANSAC~\cite{loransac2003} on $2$-D to $3$-D correspondences, and then refining the pose of the new image by minimizing its re-projection error. $\mathsf{ii-Triangulation:}$ scene points of the new image are triangulated and added to the existing scene. $\mathsf{iii-Bundle \ Adjustment (BA):}$ this step jointly refines the camera pose and $3$-D point cloud by minimizing the total re-projection error of the currently registered images. 

Under small-parallax settings, initialization struggles to produce good initial two-view reconstruction due to unstable epipolar geometry, while incremental reconstruction tends to coverage to bad solutions due to large triangulation variation. We now show how these two steps can benefit from depth-prior obtained from a pretrained network.
Note that we do not modify BA as our improved previous steps already provide a strong starting point where adding depth-prior to BA does not result in any additional gains.
%Note that we do not modify BA since it has a sufficiently good starting point with improved previous steps and adding depth-prior to it does not bring additional gains.

\subsection{Finding Keypoint-Depth}

\noindent Given an image-set, we use standard COLMAP~\cite{colmap2016} pipeline for $2$-D keypoints detection and matching. Moreover, we use a pretrained monocular depth-estimator to predict the dense depth map $\textbf{D}_i$ for each image $\textbf{I}_i$. The depth of keypoint $\mathbf{p}_{i}$ in $\textbf{I}_i$ is extracted from $\textbf{D}_i$ using bilinear interpolation as $\textbf{D}_{i}[\mathbf{p}_{i}]$. We incorporate this keypoint-depth in the initialization step to get a more accurate estimate of the initial camera pose and $3$-D point cloud, and regularize the optimization process of image registration and triangulation to guide the incremental reconstruction towards a better solution. Using the sparse keypoints-depth instead of dense depth map is important to maintain computation and memory efficiency for large scale reconstruction. We empirically demonstrate that our method is agnostic to the choice of depth estimation model (see $\S$~\ref{ablation}).

\subsection{Initialization}

\noindent Instead of computing the essential matrix from $2$-D to $2$-D correspondences between the initial image-pair $(\mathbf{I}_a$, $\mathbf{I}_b)$, and decomposing it into rotation and translation matrices as done in COLMAP~\cite{colmap2016}, we incorporate keypoint-depth information to formulate the initialization step as a Perspective-n-Point (PnP) problem. Specifically, we first create an initial point cloud $\mathbf{P}_a$ by projecting the $2$-D keypoints in $\mathbf{I}_a$ into $3$-D as follows:
\begin{equation}
  \mathbf{P}_a = \mathbf{D}_a[\mathbf{p}_a] \ \mathbf{K}_a^{-1} \ \mathbf{h}(\mathbf{p}_a) \quad \forall \ \ \ \mathbf{p}_a \in \mathcal{T}_a
\end{equation}

\noindent where $\mathbf{D}_a[\mathbf{p}_a]$ is the depth of $\mathbf{p}_a$, $\mathbf{K}_a \in \mathbb{R}^{3 \times 3}$ is the intrinsic matrix of the camera that captured $\mathbf{I}_a$, $\mathbf{h}(\cdot)$ converts euclidean coordinates to homogeneous coordinates, and $\mathcal{T}_a$ is the set of $2$-D keypoints in $\mathbf{I}_a$. This gives us an initial $3$-D point cloud created from keypoints in $\mathbf{I}_a$. 

The relative pose between $\mathbf{I}_a$ and $\mathbf{I}_b$ is then estimated using geometric relationship between $2$-D keypoints in $\mathbf{I}_b$ and their corresponding $3$-D points in the point cloud ($2$-D to $3$-D correspondences), which is exactly the goal of the PnP problem. Instead of estimating the relative pose using $2$-D to $2$-D correspondences with epipolar geometry which is unstable under small baseline, using $2$-D to $3$-D correspondences with PnP approach makes our initialization method much more robust to small baseline since PnP naturally prefers small baseline data.

Note that unlike COLMAP~\cite{colmap2016} which selects the initial image-pair by considering both triangulation angle and the number of matched keypoints, we select the image pair which has the largest number of matched keypoints with valid depth. We consider depth to be valid for all values except $0$ or infinity. Due to our large range of acceptable depth, more matched keypoints are used to generate the initial point cloud with larger scene-coverage, making subsequent reconstruction steps more robust and accurate.

\subsection{Depth-Regularized Optimization}

\begin{figure}[!t]
	\centering
	\includegraphics[width=0.375\textwidth]{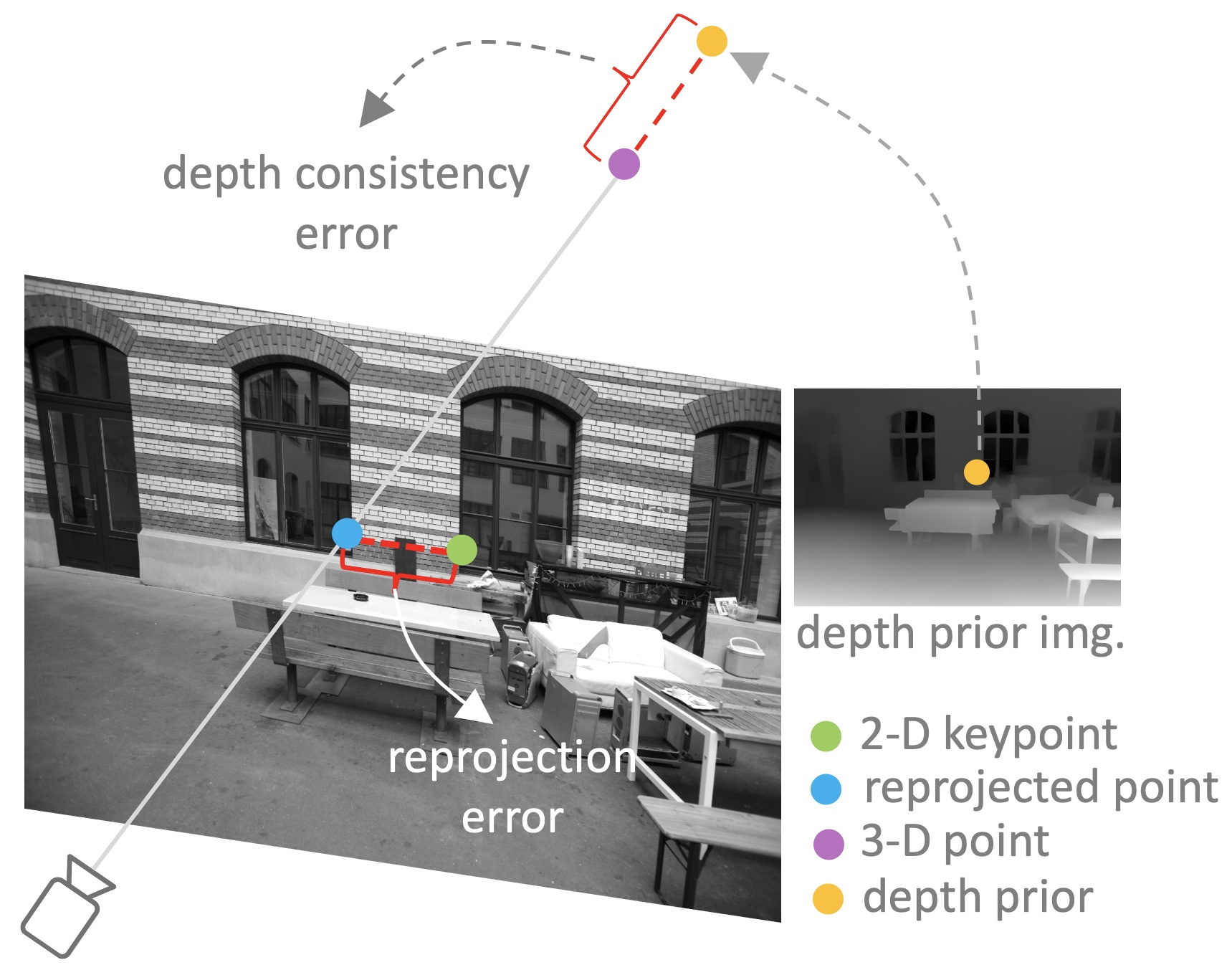}
\vspace{-0.1cm}	\caption{\small{\textbf{Depth Regularized Optimization -- } In addition to the generally used re-projection error, we use depth-consistency error as a regularizer for optimizing our loss functions for image registration (Equation~\ref{eq:reg_error}) as well as triangulation (Equation~\ref{eq:triangularion_error}).}}\vspace{-0.2cm}
	\label{fig:depth_consistency}
\end{figure}

\noindent The initialization step is followed by: (a) image registration, which registers a new image to the existing scene and (b) triangulation, which triangulates the new points. We define a novel depth-regularized objective to improve these two steps. The intuition of our approach is illustrated in Figure \ref{fig:depth_consistency} and its details are explained below.

\vspace{0.1cm} \noindent \textbf{a. Image Registration:} We follow the procedure used in COLMAP~\cite{colmap2016} to select our next image $\textbf{I}_i$, and estimate its initial camera pose using PnP problem formulation with RANSAC~\cite{loransac2003}. We further refine this initial camera pose by minimizing the following objective function:
\begin{equation}
  \textbf{R}_{i}^{*}, \textbf{t}_{i}^{*}, \gamma_i, \beta_i = \arg\min_{\textbf{R}, \textbf{t}, \gamma, \beta} \sum_{\mathbf{p}_{i} \in \varphi_{i}} \mathrm{E}_{\textrm{PR}}(\textbf{p}_{i}) + \lambda \mathrm{E}_{\textrm{DC}}(\textbf{p}_{i}, \gamma_{i}, \beta_{i})
  \label{eq:reg_error}
\end{equation}
\noindent Here, $\varphi_{i}$ is the set of inliers keypoints obtained from RANSAC of initial pose estimation, while $\mathrm{E}_{\mathrm{PR}}$ is the reprojection loss and $\mathrm{E}_{\mathrm{DC}}$ is the depth consistency loss. $\lambda$ is the weight to balance the two losses. $\mathrm{E}_{\mathrm{PR}}$ is defined as:
\begin{equation}
  \mathrm{E}_{\textrm{PR}}(\textbf{p}_{i}) = ||\Pi(\mathbf{R}_i \mathbf{P}_{i} + \mathbf{t}_i) - \textbf{p}_{i}||
\end{equation}
\noindent where $\Pi$ represents the projection from $3$-D points to image plane. Similarly $\mathrm{E}_{\textrm{DC}}$ is defined as:
\begin{equation}
  \mathrm{E}_{\textrm{DC}}(\mathbf{p}_{i}, \gamma_i, \beta_i) = \sum_{\mathbf{p}_{i} \in \varphi(i)}||[\mathbf{R}_i \mathbf{P}_{i} + \mathbf{t}_i]_{z} - \gamma_i \textbf{D}_i[\textbf{p}_{i}] - \beta_i||
\end{equation}
\noindent where $\textbf{D}_i[\mathbf{p}_{i}]$ is the depth of keypoints $\mathbf{p}_{i}$, $[\mathbf{x}]_{z} \in \mathbb{R} (\mathbf{x} \in \mathbb{R}^{3})$ is the third element of the $3$-D point $\mathbf{x}$. $\gamma_i$ and $\beta_i$ are the scale and shift to align the depth-prior of $\textbf{I}_i$ with the projected depth from $3$-D points..

\vspace{0.1cm} \noindent \textbf{b. Triangulation: } Once image $\mathbf{I}_i$ is registered, the newly observed scene points are added to the existing point cloud via triangulation. We first use DLT~\cite{dlt_2004} and RANSAC~\cite{loransac2003} to estimate the initial $3$-D position and refine it using the following objective function:
\begin{equation}
\mathbf{P}_{i}^{*} = \arg\min_{\mathbf{P}_{i}^{\textrm{new}}} \sum_{\mathbf{p}_j \in \textrm{N}(\mathbf{P}_{i}^{\textrm{new}})}\textrm{E}_{\textrm{PR}}(\mathbf{p}_j) + \lambda \textrm{E}_{\textrm{DC}}(\mathbf{p}_j, \gamma_i, \beta_i)
\label{eq:triangularion_error}
\end{equation}
\noindent where $\mathbf{P}_{i}^{\textrm{new}}$ is the new set of $3$-D points observed in $\mathbf{I}_{i}$, $\textrm{N}(\mathbf{P}_{i}^{\textrm{new}})$ is the set of $2$-D keypoints corresponding to $\mathbf{P}_{i}^{\textrm{new}}$, and $\textrm{E}_{\textrm{PR}}$ and $\textrm{E}_{\textrm{DC}}$ are the reprojection and depth consistency errors as defined above. $\gamma_i$ and $\beta_i$ are computed from image registration and are kept fixed here.
Note that the $3$-D point estimated from triangulation based only on reprojection loss has large variance when the triangulation angle is small~\cite{lee2019triangulation}. Our objective function addresses this challenge by regularizing the position of the $3$-D point using the depth consistency error while keeping the reprojection error low.

\section{Experiments}

\subsection{Datasets}
\label{ss:datasets}
\noindent We first go over the datasets we used in our experiments.

\vspace{0.1cm} \noindent \textbf{a. StudioSfM:} To undertake a comprehensive comparative evaluation of our approach on studio-produced video-content, we collected a new dataset called StudioSfM which contains $130$ shots with $21\textrm{K}$ frames from $15$ TV-episodes. For each full-length TV episode, we first ran shot segmentation~\cite{shot2011} to split it into a set of constituent shots and then sparsely sampled these shots in a uniform manner. For each sampled shot, we let professional visual-effects artists generate the ground-truth camera poses and $3$-D point clouds through commercial CG software by manually tracking high-quality features, identifying co-planner constraints, and adjusting focal length.
We removed the shots which were too challenging to be annotated due to factors such as heavy motion-blur and fully static camera.

To underscore the prevalence of small-baseline in studio-produced video-content, we compare parallax-distribution between StudioSfM dataset with a standard large-scale SfM dataset of ETH3D~\cite{eth3d2017} (shown in Figure~\ref{fig:parallax}). We computed parallax as the ratio between the maximum translation of camera motion and the median distance of $3$-D point cloud to all cameras. Figure~\ref{fig:parallax} shows that most videos of StudioSfM have small parallax because the shots in movies and TV shows tend to have less camera motion to create an engaging viewing experience. In contrast, ETH3D has a much larger parallax since it is captured specifically for the purposes of $3$-D reconstruction using standard approaches.

\vspace{0.1cm} \noindent \textbf{b. ETH3D:} To demonstrate that our approach does not result in any accuracy loss for data with large parallax, we present experiments on ETH3D~\cite{eth3d2017} which is a standard SfM dataset and contains two categories: (a) high-res multi-view with $13$ scenes (b) low-res many-view with $5$ scenes. The precise camera poses and dense point cloud from a laser scan are provided in the dataset.

%\begin{figure}[!t]
%	\centering
%	\includegraphics[width=0.3\textwidth]{figures/parallax_comp.png}
%\vspace{-0.2cm} \caption{\small {\textbf{Parallax-Distribution Comparison:} The parallax-distribution of ETH3D~\cite{eth3d2017} is plotted with StudioSfM -- a new dataset with $16\textrm{K}$ frames in $133$ manually annotated shots from $15$ TV episodes. The long-tail distribution of StudioSfM shows that small-motion parallax is significantly more common in movies and TV shows than in standard SfM datasets.}}
%	\label{fig:parallax}
%	\vspace{-0.2cm} 
%\end{figure}

\subsection{Implementation Details}
\noindent Our approach builds on the codebase of COLMAP~\cite{colmap2016}. We use DPT-large~\cite{dptlarge2021} as our default depth estimator for producing depth-priors. The influence of using different depth models on our method is analyzed in $\S$~\ref{ablation}. We resize input image height to $384$ while maintaining the original aspect ratio. The dense depth map is resized to the original-image size using nearest neighbor interpolation. The weight $\lambda$ for depth regularized optimization is always kept fixed at $6$. Mask-RCNN~\cite{rcnn2017} is used to create binary masks of humans which are used as input for all compared approaches.

\begin{figure*}[!t]
  \centering
  \includegraphics[trim={0, 0pt, 0pt, 0pt}, clip, width=0.80\linewidth]{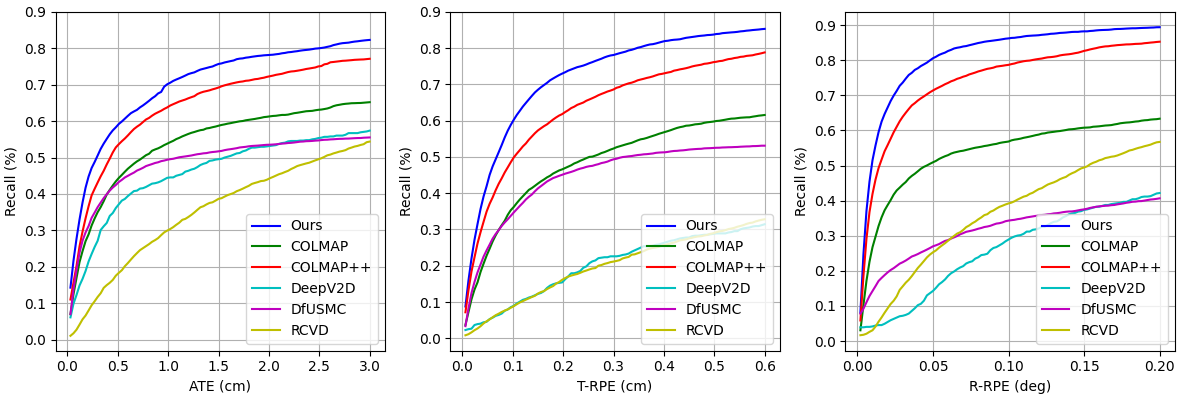}
  \vspace{-0.2cm}\caption{\small \textbf{Evaluation of camera pose on StudioSfM --} The figure shows the recall-curves of multiple comparative methods for ATE (absolute trajectory error), T-RPE (relative pose error for translation) and R-RPE (relative pose error for rotation).}\vspace{-0.2cm}
  \label{fig:pose_all}
\end{figure*}

\subsection{Baselines}

\noindent Unlike comparisons provided by previous approaches ~\cite{rcvd2021, Lai21a} which only use original COLMAP~\cite{colmap2016} on videos with small camera-motion, we fine-tune its hyper-parameters for small-parallax setting to make it much less likely to fail on small-parallax data in order to have a more fair comparison. We call this version of COLMAP~\cite{colmap2016} as COLMAP++, and compare our approach against the original COLMAP\cite{colmap2016}, COLMAP++, DeepV$2$D~\cite{deepv2d2020}, RCVD~\cite{rcvd2021} and DfUSMC~\cite{small2016}.

\subsection{Evaluation Metrics}

\noindent To evaluate camera pose, we compute three commonly used metrics: absolute trajectory error (ATE), relative pose error for translation (T-RPE) and rotation (R-RPE). We refer to the work of \cite{prokhorov2019measuring} for detailed explanation of these metrics. To evaluate $3$-D point cloud, we project point cloud to each frame using estimated camera-pose and measure the accuracy of relative depth $\delta=\textrm{max}(\frac{y_i^*}{y_i}, \frac{y_i}{y_i^*})$ and absolute depth $\theta=|y_i - y_i^*|$ under different thresholds where $y_i$ and $y_i^*$ are the estimated and ground truth depth respectively.

\begin{table}[b]
	\Huge
  \centering
  \resizebox{\columnwidth}{!} {
  \begin{tabular}{|l||c|c|c|c|c|c|}
    %\toprule
    \hline
    \multirow{2}{*}{Method} & \multicolumn{2}{c|}{ATE AUC} & \multicolumn{2}{c|}{T-RPE AUC} & \multicolumn{2}{c|}{R-RPE AUC} \\
    % \cmidrule(lr){2-5}
    % \cmidrule(lr){6-9}
    % \cmidrule(lr){10-13}
    %\cmidrule(lr){2-3}
    %\cmidrule(lr){4-5}
    %\cmidrule(lr){6-7}
    \cline{2-7}
    % \cmidrule(lr){8-9}
    % \cmidrule(lr){10-11}
    % \cmidrule(lr){12-13}
    & 0.2 (cm) & 2.0 (cm) & 0.1 (cm) & 0.5 (cm) & 0.02 ($^{\circ}$) & 0.1 ($^{\circ}$) \\
    \hline\hline
    % DeepV2D~\cite{teed2018deepv2d} \\
    % RCVD~\cite{kopf2021robust} \\
    % DFUSMC~\cite{ha2016high} \\
    RCVD~\cite{rcvd2021} & 4.2 & 28.0 & 5.0 & 17.5 & 4.4 & 23.1 \\
    DfUSMC~\cite{small2016} & 22.1 & 45.8 & 23.8 & 43.0 & 14.5 & 25.7 \\
    DeepV2D~\cite{deepv2d2020} & 15.2 & 43.8 & 5.6 & 19.2 & 4.6 & 15.8 \\
    COLMAP~\cite{colmap2016} & 20.0 & 49.6 & 23.1 & 45.8 & 25.8 & 46.7\\ 
    COLMAP++ & 24.7 & 59.1 & 34.3 & 60.1 & 39.7 & 66.0\\
    \hline
    Ours & \textbf{31.8}{\color{Green}+7.1} & \textbf{65.3}{\color{Green}+6.2} & \textbf{41.6} {\color{Green}+7.3} & \textbf{69.8} {\color{Green}+9.7} & \textbf{48.7} {\color{Green}+9.0}& \textbf{74.8} {\color{Green}+8.8}\\
    \bottomrule
  \end{tabular}
  }
  \vspace{-0.2cm}\caption{\small \textbf{Camera pose evaluation on StudioSfM using AUC --}. Recall-curve AUCs for our three considered metrics are shown.}\vspace{-0.3cm}
\label{pose_auc}
\end{table}

\subsection{Results}
\subsubsection{StudioSfM}
\noindent \textbf{a. Camera Pose Evaluation:} We first evaluate the quality of the estimated camera pose on StudioSfM dataset. The predicted camera poses are aligned with the ground truth camera pose using similarity transformation before computing the metrics. Figure \ref{fig:pose_all} shows the plot of recall against three error metrics and Table \ref{pose_auc} shows the area-under-curve (AUC) for each curve. Our approach significantly outperforms other approaches across all three metrics. COLMAP++ performs much better than original COLMAP~\cite{colmap2016} showing the importance of tuning it to work with small parallax datasets. DfUSMC~\cite{small2016} does not work well on StudioSfM which indicates that their assumptions about camera-motion do not generalize to our data. DeepV2D~\cite{deepv2d2020} also shows low performance on StudioSfM likely due to their lack of outliers handling mechanisms.

\begin{figure}[!t]
	\centering
	\includegraphics[height = 0.85\linewidth]{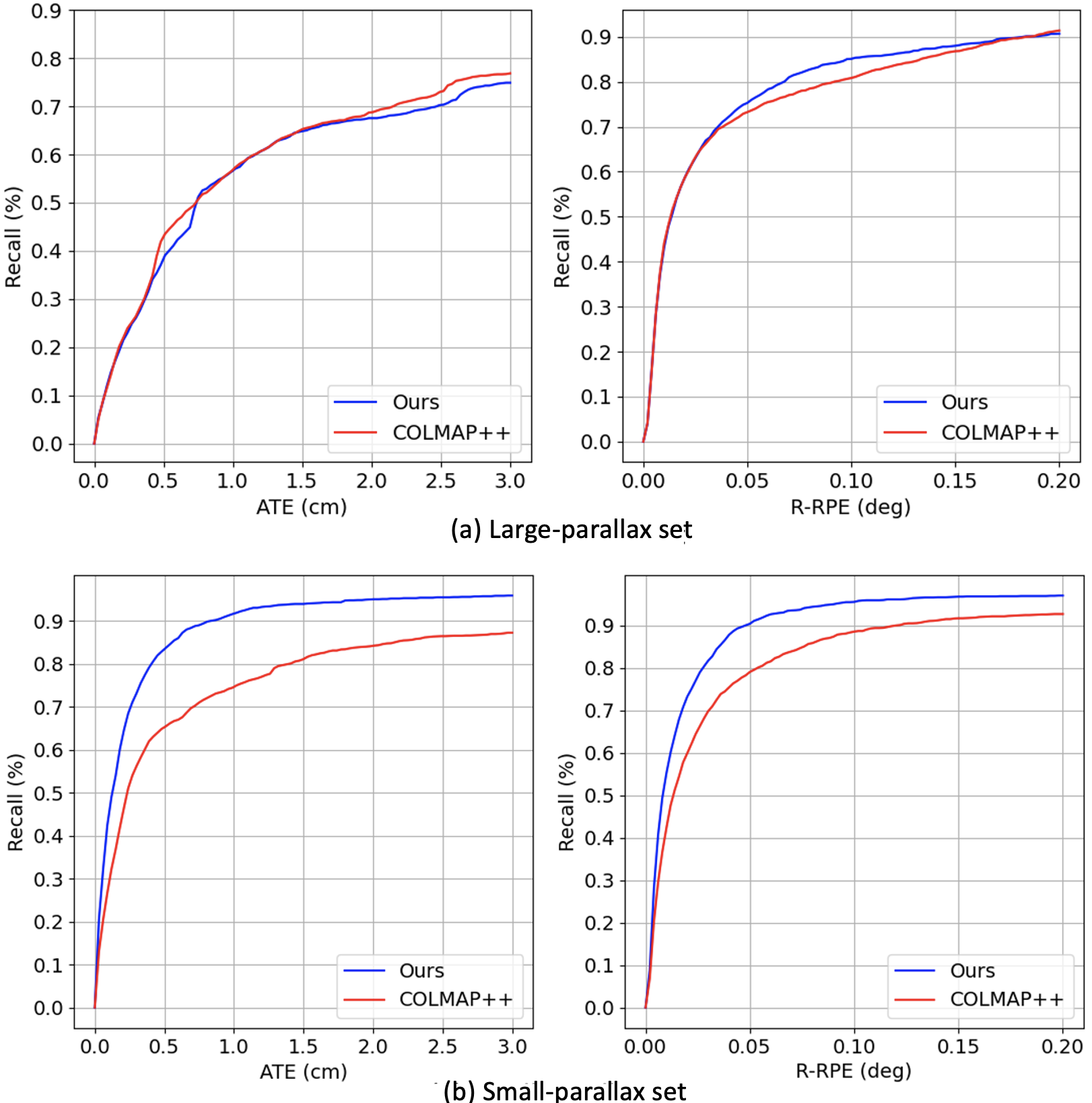}
	\caption{\small \textbf{Camera pose estimation using StudioSfM under:} (a) large-parallax set and (b) small-parallax set. Our approach offers significant improvement over COLMAP++ on small-parallax data.}\vspace{-0.2cm}
	\label{fig:parallax_set}
\end{figure}

To further clarify the benefit of our approach for small-parallax settings, we sort videos in StudioSfM data according to their parallax in descending order, and use the top $30$\% of data as the large-parallax set and bottom $30$\% as the small-parallax set. We compare our estimated camera pose with COLMAP++ using these two sets. Figure~\ref{fig:parallax_set} shows the significantly better performance of our approach on small-parallax set, highlighting the importance of using depth-priors in geometry-based SfM for small-parallax settings.

%Figure~\ref{fig:parallax_set} shows that our approach has significantly better performance than COLMAP++ on the small-parallax set. This highlights the importance of using depth-priors in geometry-based SfM pipelines for small-parallax settings.

%The significantly better results of our approach shown in Figure~\ref{fig:parallax_set} highlight the importance of using depth-priors in geometry-based SfM pipelines for small-parallax settings.

%Figure~\ref{fig:parallax_set} shows the significantly better results of our approach on small-parallax set highlighting the importance of using depth-priors in geometry-based SfM pipelines for small-parallax settings.

\vspace{0.1cm} \noindent \textbf{b. Point Cloud Evaluation:} To evaluate the quality of estimated point clouds, we first project point clouds into each frame using the estimated camera-pose and then compare depths of the projected points with ground truth depths. Besides computing the accuracy using relative depth error as done in DeepV2D~\cite{deepv2d2020}, we also compare the accuracy using absolute depth error as our ground truth point clouds are annotated using real-world scale. Table~\ref{table: pointcloud} shows that our approach outperforms all other approaches on both relative and absolute depth error. Directly applying DPT-large~\cite{dptlarge2021} does not produce accurate depths even though they can visually look good. In contrast, our method of using of the output of DPT-large~\cite{dptlarge2021} as depth-priors in geometry-based SfM substantially improves the quality of estimated depth.

\begin{table}
\Huge
\begin{center}
\resizebox{\columnwidth}{!} {
\begin{tabular}{|l||c|c|c|c|c|c|}
\hline
Method & \multicolumn{3}{c|}{Relative depth accuracy (\%)} & \multicolumn{3}{c|}{Absolute depth accuracy (\%)} \\
\cline{2-7}
 & $\delta < 1.25 $ & $\delta < 1.25^2$ & $\delta < 1.25^3$ & $\theta < $ 5cm & $\theta < $ 10cm & $\theta < $ 25cm \\
\hline\hline
$\text{DPT-large}^*$ \cite{dptlarge2021} & 33.5 & 53.6 & 64.9 & 3.8 & 7.4 & 14.5 \\
RCVD~\cite{rcvd2021} & 43.7 & 66.5 & 79.4 & 5.2 & 9.1 & 18.7\\
DfUSMC~\cite{small2016} & 27.6 & 39.6 & 46.4 & 2.4 & 4.5 & 8.9\\
DeepV2D~\cite{deepv2d2020} & 63.4 & 80.1 & 87.5 & 8.9 & 15.5 & 28.8\\
COLMAP~\cite{colmap2016} & 50.8 & 55.1 & 56.9 & 20.7 & 27.3 & 38.3\\
COLMAP++ & 72.9 & 81.4 & 85.0 & 22.8 & 32.6 & 50.6 \\
\hline
Ours & \textbf{80.0} & \textbf{86.0} & \textbf{89.3} & \textbf{27.1} & \textbf{39.0} & \textbf{57.3} \\
\hline
\end{tabular}
}
\end{center}
\vspace{-0.5cm}\caption{\small \textbf{Evaluation of depth estimation on StudioSfM --} Recall is measured here using relative depth and absolute depth. Results for DPT-large~\cite{dptlarge2021} are presented here as a reference.}\vspace{-0.2cm}
\label{table: pointcloud}
\end{table}

\begin{table}
	\Huge
  \centering
  \resizebox{\columnwidth}{!} {
  \begin{tabular}{|l||c|c|c|c|c|c|}
  	\hline
    %\toprule
    \multirow{2}{*}{Method} & \multicolumn{2}{c|}{ATE AUC} & \multicolumn{2}{c|}{T-RPE AUC} & \multicolumn{2}{c|}{R-RPE AUC} \\
    %\cmidrule(lr){2-3}
    %\cmidrule(lr){4-5}
    %\cmidrule(lr){6-7}
    \cline{2-7}
    & 0.2 (cm) & 2.0 (cm) & 0.1 (cm) & 0.5 (cm) & 0.02 ($^{\circ}$) & 0.1 ($^{\circ}$) \\
    \midrule
	& \multicolumn{6}{c|}{high-res multi-view}\\
    \hline
    COLMAP~\cite{colmap2016} & 95.7 & 99.4 & 96.7 & 98.7 & 27.2 & \textbf{70.6}\\
    \hline
    Ours & \textbf{99.5} & \textbf{99.9} & \textbf{97.1} & \textbf{99.1} & \textbf{27.7} & 69.8\\
    \midrule
    & \multicolumn{6}{c|}{low-res many-view}\\
    \hline
    COLMAP~\cite{colmap2016} & 18.6 & 74.3 & 65.8 & 92.5 & \textbf{0.5} & 7.5 \\
    \hline
    Ours & \textbf{42.1} & \textbf{88.8} & \textbf{86.4} & \textbf{96.9} & 0.4 & \textbf{14.8}\\
    \bottomrule
  \end{tabular}
  }
 \caption{\small \textbf{Camera pose evaluation on the two categories of ETH3D using AUC --} We report AUC of each metric. Our approach achieves results comparable with COLMAP on high-res multi-view category for all metrics, and outperforms COLMAP on low-res many-view category for most metrics.}\vspace{-0.2cm}
\label{pose_auc_eth}
\end{table}

\subsubsection{ETH3D}
To demonstrate the effectiveness of our approach on standard SfM datasets, we assess it on two categories of ETH3D~\cite{eth3d2017} where motion-parallax is significantly larger than StudioSfM. Our approach is compared with original COLMAP~\cite{colmap2016} which is already tuned for large-parallax. The camera pose comparison is presented in Table~\ref{pose_auc_eth}. On high-res multi-view category both COLMAP~\cite{colmap2016} and our method achieve impressive performance while our method is still able to slightly outperform COLMAP~\cite{colmap2016}. Our clear gains over COLMAP~\cite{colmap2016} on low-res many-view category show that our approach is more robust to low resolution images than COLMAP~\cite{colmap2016}. The comparison of estimated depth using high-res multi-view category is shown in Table~\ref{eth_depth} in which we achieve better absolute depth accuracy than COLMAP~\cite{colmap2016}. Our overall better performance on ETH3D demonstrates that our approach does not show any degradation on large-parallax data while offering significant gains for small-parallax settings.

\begin{table}
\Huge
\begin{center}
\resizebox{\columnwidth}{!} {
\begin{tabular}{|l||c|c|c|c|c|c|}
\hline
Method & \multicolumn{3}{c|}{Relative depth accuracy (\%)} & \multicolumn{3}{c|}{Absolute depth accuracy (\%)} \\
\cline{2-7}
 & $\delta < 1.05 $ & $\delta < 1.05^2$ & $\delta < 1.05^3$ & $\theta < $ 1cm & $\theta < $ 2cm & $\theta < $ 5cm \\
\hline\hline
COLMAP~\cite{colmap2016} & \textbf{96.9} & \textbf{98.1} & \textbf{98.5} & 58.7 & 72.9 & 86.0\\
\hline
Ours & 96.8 & 98.0 & 98.4 & \textbf{61.2} & \textbf{75.7} & \textbf{88.1} \\
\hline
\end{tabular}
}
\end{center}
\vspace{-0.4cm}\caption{\small \textbf{Evaluation of depth estimation on ETH3D high-res multi-view category --} Accuracy is measured here using both relative depth and absolute depth.}\vspace{-0.2cm}
\label{eth_depth}
\end{table}

\subsection{Ablation Study} \label{ablation}

\begin{figure}[!t]
	\centering
	\includegraphics[width=0.475\textwidth, height=0.22\textwidth]{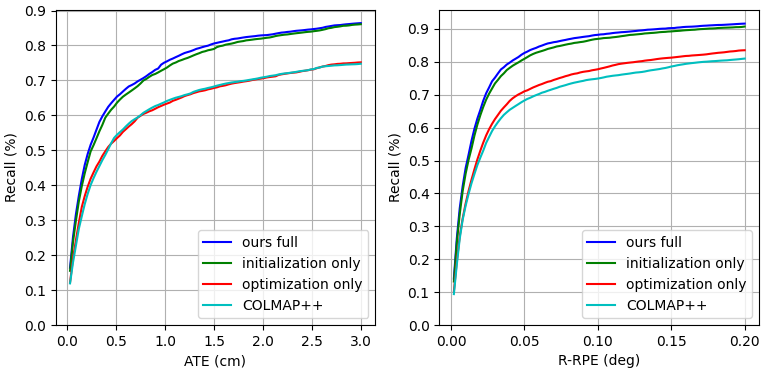}
	\caption{\small \textbf{Ablation study on StudioSfM --} Recall of translation error and relative rotation error are plotted for different variants of our method: "ours full" - our full approach, "initalization only" - our approach with only improved initialization, "optimization only" - our approach with only depth-regularized optimization.}\vspace{-0.2cm}
	\label{fig:variants}
\end{figure}

\vspace{0.1cm} \noindent \textbf{a. Method Variants:} We compare several variants of our approach with COLMAP++ on StudioSfM dataset. Figure \ref{fig:variants} compares the recall curves for ATE and R-RPE between COLMAP++, our approach with only improved initialization (initialization only), our approach with only depth-regularized optimization (optimization only) and our full approach (ours full). We can see that our proposed initialization using depth-prior of keypoints achieves substantial improvement over COLMAP++ showing the criticality of initialization for SfM pipeline to converge to a good solution. With both improved initialization and depth regularized optimization, our full approach performs the best.

\begin{figure}[!t]
	\centering
	\includegraphics[width=0.475\textwidth, height=0.22\textwidth]{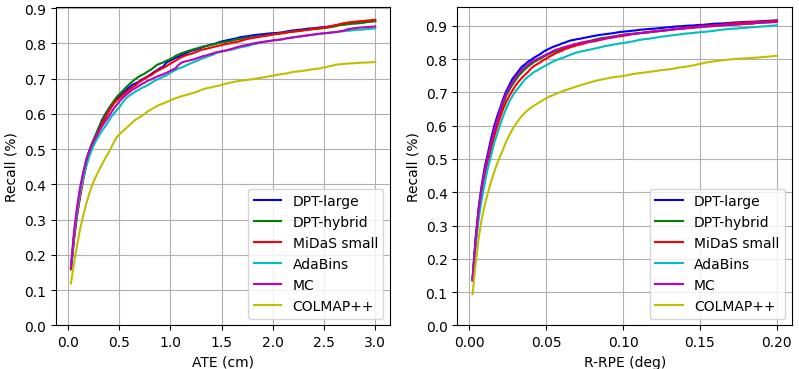}
	\caption{\small \textbf{Ablation study of depth-estimators for StudioSfM --} The recalls for ATE and R-RPE are plotted for our approach with different pretrained monocular depth estimators.}
	\vspace{-0.2cm}
	\label{fig:depths}
\end{figure}

\vspace{0.1cm} \noindent \textbf{b. Depth Estimators:} To assess the robustness of our approach to the choice of depth-estimator, we evaluate camera pose estimation using several off-the-shelf pretrained depth estimation models based on various network architectures and trained with different datasets. Specifically, we compare five monocular depth estimation models, including MiDaS small~\cite{midas2020} which is designed for mobile devices, DPT-hybrid~\cite{dptlarge2021} and DPT-large~\cite{dptlarge2021} which are based on Transformers~\cite{vaswani2017attention}, AdaBins~\cite{adabin2021} which is the latest approach for monocular depth estimation and MC~\cite{mc2019} which focuses on human depth estimation. Figure~\ref{fig:depths} shows that our approach significantly outperforms COLMAP++ using depth priors provided by any of the five different pretrained depth estimation models. The small performance variation among those depth estimators demonstrates that our approach does not rely on a particular depth estimator and is robust to diverse network architectures and training datasets.

\vspace{0.1cm} \noindent \textbf{c. Depth Noise:} In addition to evaluating the use of various depth estimators we also test the robustness of our approach under different amounts of synthetic noise. For each keypoint depth $d$, we add random Gaussian noise with $0$ mean and $\alpha \cdot d$ standard deviation with different values of $\alpha$. As shown in Table~\ref{depth_noise}, the performance degradation of our approach is only within $5\%$ under the largest added noise level of $0.4$ which demonstrates that our pipeline can tolerant sizable amounts of errors in the estimated depth-priors.

\begin{table}
	%\Huge
  \centering
  \resizebox{\columnwidth}{!} {
  \begin{tabular}{|l||c|c|c|c|c|c|}
    \hline
    \multirow{2}{*}{$\alpha$} & \multicolumn{2}{c|}{ATE AUC} & \multicolumn{2}{c|}{T-RPE AUC} & \multicolumn{2}{c|}{R-RPE AUC} \\
    \cline{2-7}
    & 0.2 (cm) & 2.0 (cm) & 0.1 (cm) & 0.5 (cm) & 0.02 ($^{\circ}$) & 0.1 ($^{\circ}$) \\
    \hline\hline
    \textbf{0.0} & 31.8 & 65.3 & 41.6 & 69.8 & 48.7 & 74.8 \\
    \textbf{0.1} & 31.1 & 63.9 & 39.4 & 67.6 & 47.1 & 73.0 \\
    \textbf{0.2} & 30.3 & 62.8 & 39.2 & 67.5 & 47.0 & 73.5 \\
    \textbf{0.4} & 28.5 & 62.7 & 38.5 & 67.0 & 44.6 & 71.9 \\
    \bottomrule
  \end{tabular}
  }
  \vspace{-0.2cm}\caption{\small \textbf{Depth noise analysis for camera pose estimation -- }. 
  Gaussian noise with $0$ mean and $\alpha \cdot d$ variance is added to each keypoint with depth $d$. 
  The performance degradation of our approach is only within $5\%$ for $\alpha$ of $0.4$ demonstrating the robustness of our approach to errors in estimated depth-priors.
  }\vspace{-0.3cm}
\label{depth_noise}
\end{table}

\begin{figure}[!t]
  \centering
  \includegraphics[trim={0, 0pt, 0pt, 0pt}, clip, width=1.00\linewidth]{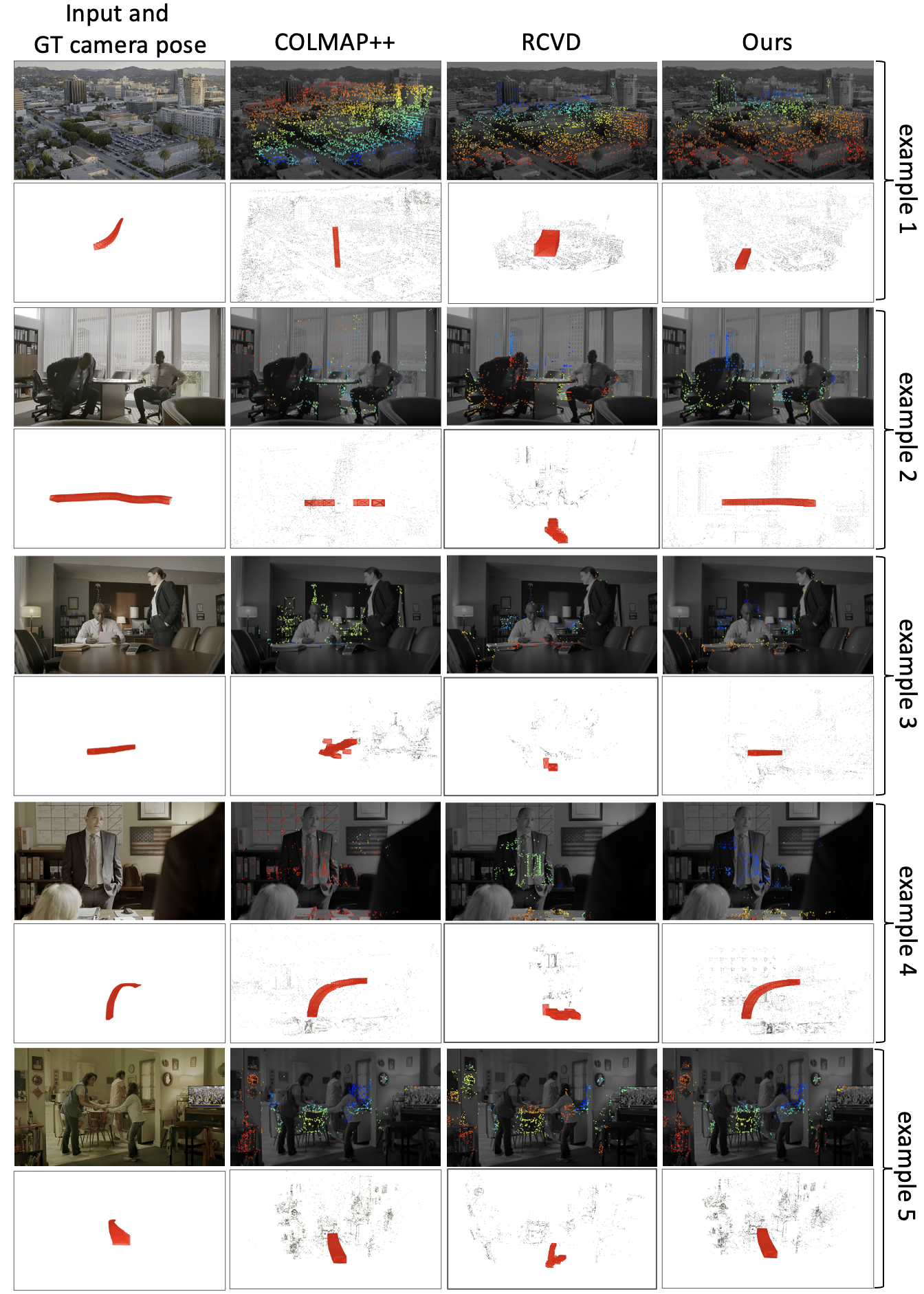}
  \caption{\small \textbf{Qualitative results on StudioSfM dataset --} Keypoint-depths are visualized in color from {\color{Red} red} (near) to {\color{Blue} blue} (far) and the camera motion is visualized as a trajectory of red cones. 
  First column shows image and ground truth camera motion, while other columns show the results from different approaches.}
  \vspace{-0.2cm}
  \label{fig:examples}
\end{figure}

\subsection{Qualitative Evaluation}

\vspace{0.1cm} \noindent \textbf{a. StudioSfM:} We compare our approach with other methods qualitatively using five examples from StudioSfM in Figure \ref{fig:examples}. To compare with RCVD~\cite{rcvd2021}, we use their estimated depth image to visualize the depth of the point cloud. Examples $1$-$4$ show a common error observed for COLMAP++ where, unlike our approach, the relative depths between points are incorrect (\textit{e.g.}, the building outside of window in example $2$ is estimated closer than the table in the room). Similarly, the camera motion estimated by RCVD~\cite{rcvd2021} tends to have large errors as shown in example $2$-$5$. Both COLMAP++ and our approach achieve accurate reconstruction for example $5$ since the motion-parallax is sufficient, however, RCVD~\cite{rcvd2021} still produces poor results for this example even though the motion parallax is large.

\begin{figure}[!t]
  \centering
  \includegraphics[trim={0, 0pt, 0pt, 0pt}, clip, width=1.00\linewidth]{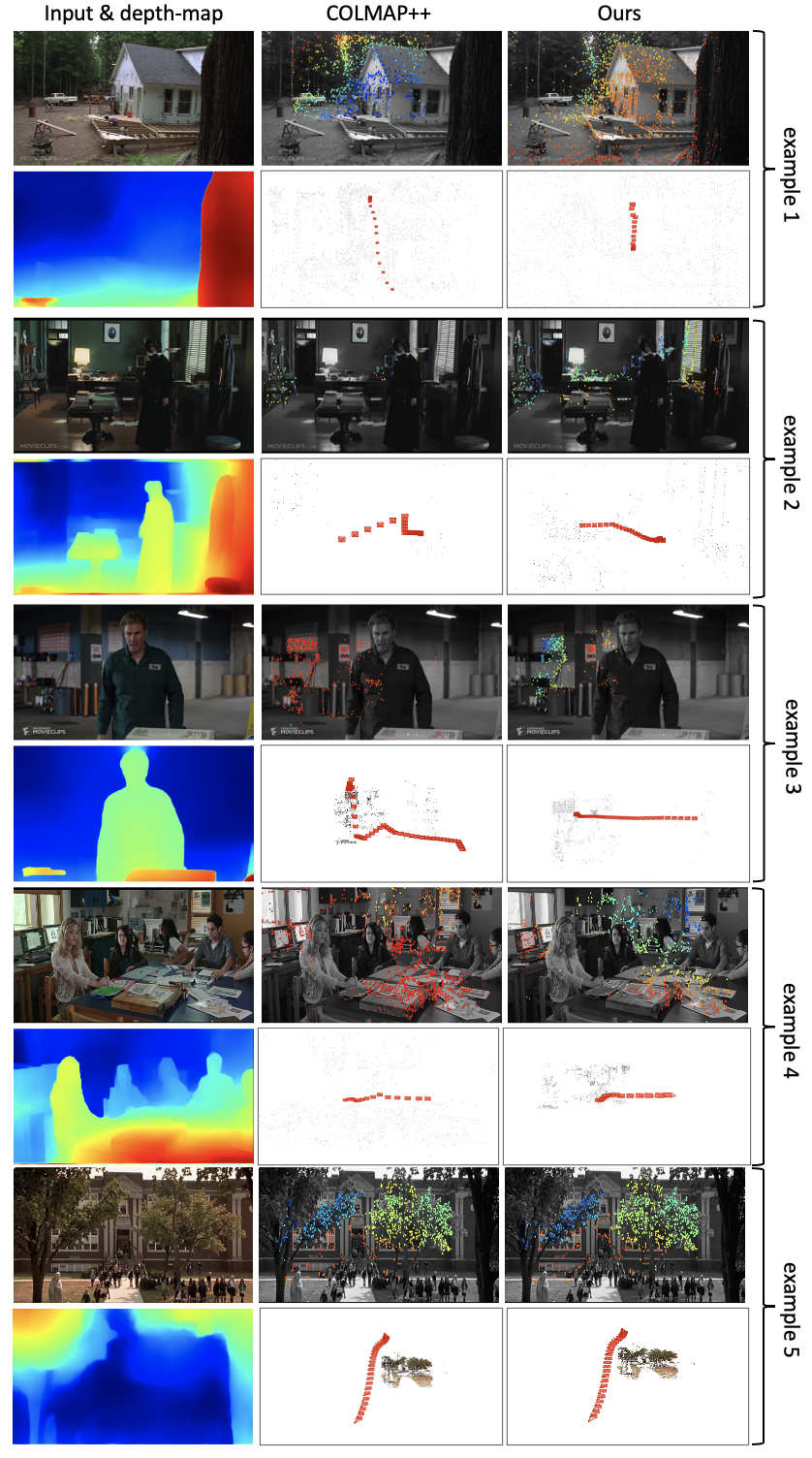}
  \caption{\small \textbf{Qualitative results on LVU dataset~\cite{lvu2021} -- } Depths are visualized using {\color{Red} red} (near) to {\color{Blue} blue} (far). $1^{\textrm{st}}$ column shows the input image and its depth, while $2^{\textrm{nd}}$ and $3^{\textrm{rd}}$ columns show the results of COLMAP++ and our approach. }
 % The first four %rows present examples where our approach produces better results than COLMAP++, while the last row shows an example where both approaches produce inaccurate reconstructions.}
  \vspace{-0.2cm}
  \label{fig:examples_lvu}
\end{figure}

\vspace{0.1cm} \noindent \textbf{b. LVU Dataset:} We now present qualitative results of our approach and COLMAP++ on a subset of the LVU dataset~\cite{lvu2021} consisting of video clips from movies. We selected $53$ shots with relatively few dynamic objects and small motion blur from the test-set of category "scene". As there is no ground truth provided, we can only evaluate the results by visualizing the camera poses and point clouds. Out of the selected $53$ shots, we did not find any shot where results from COLMAP++ were clearly better than ours. Figure \ref{fig:examples_lvu} shows results of $5$ examples demonstrating the higher-quality results produced by our approach. The last row shows an example where our approach produces similar errors as COLMAP++. This is because the estimated depth images~\cite{dptlarge2021} of initial image-pair for this example are too erroneous for our approach to effectively guide the subsequent reconstruction process to a better solution.

\section{Conclusions}
\noindent We presented a simple yet effective SfM approach that uses monocular depth obtained from a pretrained network to improve the incremental SfM pipeline~\cite{colmap2016}.
Experiments using existing and a newly collected dataset show that our approach significantly improves the reconstruction quality for small parallax data while being robust to a variety of pretrained depth networks. Our approach easily integrates with COLMAP~\cite{colmap2016}, and going forward we plan to use it as an initial step for dense reconstruction and novel view synthesis for studio-produced content.

%The smooth integration with COLMAP~\cite{colmap2016} makes our approach easily to be used as the preliminary step for researches such as dense reconstruction or novel view synthesis for studio-level content.

%Going forward, we will further improve the robustness of our approach to more effectively deal with situations where  depth-priors might have significantly higher amounts of errors.

%%%%%%%%% REFERENCES
{\small
\bibliographystyle{ieee_fullname}
\bibliography{egbib}
}

\end{document}